\documentclass[a4paper]{article}

\usepackage[english]{babel}
\usepackage[utf8x]{inputenc}
\usepackage[T1]{fontenc}
\usepackage{multirow}
\usepackage{authblk}
\usepackage[a4paper,top=3cm,bottom=2cm,left=3cm,right=3cm,marginparwidth=1.75cm]{geometry}

\usepackage{amsmath}
\usepackage{graphicx}
\usepackage[colorinlistoftodos]{todonotes}
\usepackage[colorlinks=true, allcolors=blue]{hyperref}
\usepackage{subcaption}

\title{Automatic L3 slice detection in 3D CT images using fully-convolutional networks}

\author{Fahdi Kanavati}
\author{Shah Islam}
\author{Eric O. Aboagye}
\author{Andrea Rockall}
\affil{Comprehensive Cancer Imaging Centre, Hammersmith Hospital, Imperial College London, UK}
\date{}                     %

\newcommand{\numImages}{1070 }
\newcommand{\numTrainImages}{1006 }
\newcommand{\numTransImages}{57 }
\begin{document}
\maketitle

\begin{abstract}
The analysis of single CT slices extracted at the third lumbar vertebra (L3) has garnered significant clinical interest in the past few years, in particular in regards to quantifying sarcopenia (muscle loss).
In this paper, we propose an efficient method to automatically detect the L3 slice in 3D CT images.
Our method works with images with a variety of fields of view, occlusions, and slice thicknesses.
3D CT images are first converted into 2D via Maximal Intensity Projection (MIP), reducing the dimensionality of the problem. 
The MIP images are then used as input to a 2D fully-convolutional network to predict the L3 slice locations in the form of 2D confidence maps.
In addition we propose a variant architecture with less parameters allowing 1D confidence map prediction and slightly faster prediction time without loss of accuracy.
Quantitative evaluation of our method on a dataset of \numTrainImages 3D CT images yields a median error of 1mm, similar to the inter-rater median error of 1mm obtained from two annotators, demonstrating the effectiveness of our method in efficiently and accurately detecting the L3 slice.
Code and dataset will be made available at \url{https://github.com/fk128/ct-slice-detection}.
\end{abstract}

\section{Introduction}

The analysis of CT slices extracted at the third lumbar vertebra (L3) has garnered significant clinical interest in the past few years, in particular in regards to computing a sarcopenia measure \cite{shachar2016prognostic}.
Sarcopenia refers to loss of muscle mass and is computed as the total area of the skeletal muscle mass divided by the square of the patient's height.
Sarcopenia is particularly relevant in oncology where severe muscle loss in adult patients is typically found to be associated with poor outcome \cite{tegels2015sarcopenia,mir2012sarcopenia, kazemi2016computed}.
L3 is taken as a standard landmark by a majority of medical researchers for sarcopenia measurement \cite{shachar2016prognostic}, as muscle and adipose fat areas at L3 and L4 have been found to be most linearly correlated to their whole-body counterparts \cite{kazemi2016computed}.

The main motivation for automating the whole process of computing a sarcopenia measure is to provide it as prognostic information to clinicians in cancer populations alongside the CT images.
Extracting measurement directly from CT images is convenient as CT is frequently obtained as part of cancer staging and disease assessment.
The current work-flow for computing a sarcopenia measurement is as follows:
manual extraction of the L3 slice; this involves scrolling through the 3D image slice by slice until the L3 slice is found.
Semi-automated segmentation software (e.g. Slice-O-Matic or ImageJ) that involves manual refinement is then used to segment the skeletal muscle and adipose fat tissue.
This process takes 5 to 10 minutes per image, and it becomes time-consuming to run on large datasets.

In this paper we solely address the problem of automatic slice detection, as skeletal muscle segmentation has already been addressed in the literature using fully-convolutional networks \cite{lee2017pixel, lerouge2015ioda}.
Slice detection in \cite{belharbi2017spotting} is formulated as a regression problem, where a VGG architecture with a single output fully-connected layer is used to predict the slice location.
The main advantage of the approach is that it operates on 2D images instead of 3D;
it does so by converting the 3D CT images to 2D via Maximal Intensity Projection (MIP). 
This involves projecting the maximal intensity pixel value along the perpendicular direction of the frontal plane.
The MIP image representation still contains enough information for locating the L3 vertebra.
This greatly reduces the dimensionality of the problem and allows feeding into the CNN images that have more context, as opposed to a 3D volume where input size is limited by memory capacity.
As vertebrae are similar in appearance, context is an important feature in discriminating between them.
A dense layer with a single output is attached to the last convolutional layer output of a pre-trained VGG network.
The weights are then fine-tuned.
The method only trains on image crops of fixed size ([100, 512, 3]).
However, there are a few drawbacks:
(1) In order to detect the L3 slice, a sliding window approach is adopted; this means that convolutions from overlapping image areas are being repeatedly recomputed as the window slides over the whole image, and this increases prediction time;
(2) the method has no probabilistic output as it only outputs a single number that indicates the predicted slice location;
(3) only image crops that contain the L3 vertebra are used for training, while image crops that do not contain it are excluded.
With this approach, a special elimination process is used during test time that relies on the assumption that image crops that do not contain the target vertebra will produce random outputs.
(2) and (3) could potentially be solved by adding another output to the network that serves as an indicator whether the L3 slice is present or absent.
This would allow training on negative examples and produce a probabilistic output; however, the sliding window approach is still computationally inefficient.
In this paper, we propose an efficient method to detect the L3 slice (or potentially any other slice) based on fully-convolutional networks (FCNN) that output full-resolution confidence maps either in 2D or 1D.
This is motivated by previous works \cite{tompson2015efficient, pfister2015flowing, payer2016regressing, cao2017realtime} that use FCNN to predict confidence maps for landmark localisation.

In summary, we make the following contributions: (1) an efficient method to automatically detect the L3 slice (or potentially any other slice) without using a sliding window approach; this is due to the use of a fully-convolutional network formulation for confidence map prediction; (2) we propose a novel variant architecture based on a simple modification of the UNet \cite{ronneberger2015u} architecture that produces 1D confidence map output instead of the standard 2D output, which is suited for our particular use case;  (3) we compile a large annotated dataset of \numImages images obtained from multiple public sources;
and (4) the dataset and code for reproducing our method will be publicly available online at \url{https://github.com/fk128/ct-slice-detection}.

\section{Related Work}

In computer vision, pose estimation methods such as OpenPose \cite{ cao2017realtime} and others \cite{tompson2015efficient, pfister2015flowing} rely on FCNNs to regress confidence maps of multiple landmarks, some of which achieve state-of-the-art performance.
For medical-based application, \cite{payer2016regressing} investigated the applicability of several configurations of FCNNs for multiple landmark localisation for hand X-ray images.
Notably, a UNet-based architecture \cite{ronneberger2015u}, defined by the presence of a down-sampling and up-sampling path linked via skip connections, compared to a network with a down-sampling path only, achieves the best accuracy overall for the localisation of landmarks on 2d hand X-ray image.
The up-sampling path helps in improving localisation by integrating information from higher resolution feature maps.
On the other hand, a network with only a down-sampling path outputs confidence maps of a lower resolution due to the max-pooling operations, resulting in loss of localisation precision. 
To overcome the loss in localisation precision from max-pooling, \cite{tompson2015efficient} use a 2-stage network, where the 1st stage network performs coarse localisation of a given landmark;
the output of this network, along with cropped volumes of the feature maps, from the previous layers, are used as input for a 2nd stage network to refine the location of the landmark.
This is analogous in principle to the approach adopted with a UNet-like architecture with skip connections, except that cropping is involved.
Multi-stage approaches such as \cite{tompson2015efficient, cao2017realtime} are especially helpful when the goal is to predict multiple landmarks simultaneously.
In effect, intermediate feature and confidence maps from the previous stage help in improving localisation accuracy by integrating a larger context and using the intermediate confidence maps to resolve ambiguity between landmarks of similar appearance (e.g. distinguishing between left and right arm joint).

Other works have investigated the detection of multiple vertebrae simultaneously in 3D images.
The problem is typically posed as a vertebrae detection \cite{glocker2013vertebrae, chen2015automatic, yang2017deep, yang2017automatic} and/or segmentation task \cite{ghosh2011automatic, rasoulian2013lumbar, korez2015framework}.
For machine learning-based methods, this requires ground truth annotations or segmentations to be available for all the vertebrae that appear in the images of the training set; a drawback of this is an increase of ground truth annotation time.
In \cite{glocker2013vertebrae} a random classification forest is used to localise vertebra centroids in 3D CT images.
The sparse centroid annotations are transformed into dense probabilistic labels for training.
Once probabilistic outputs have been predicted, a false positive removal strategy, which takes into account the global shape of the spine, is used to remove spurious outputs and improve the predictions.
\cite{chen2015automatic} use a hybrid approach that initially coarsely localises potential vertebrae candidates, then uses a CNN to output the final predictions based on the initial candidates.
\cite{suzani2015fast} extract feature vectors around voxel locations in an image and feed the feature vectors to a deep neural network to predict the presence or absence of a vertebra.
\cite{yang2017automatic} make use of a 3D CNN for vertebrae localisation.
While such approaches are beneficial if the goal is 3D vertabrae localisation, for our intended application of single slice detection that entail single vertebra detection, we find that working with 2D images provides good localisation accuracy and efficiency.

\section{Methods}
The system takes in as input a 3D CT volume. 
The volume is converted into a 2D image via  Maximal Intensity Projection (MIP) and is further post-processed.
The 2D MIP image is used as input to the network.
Depending on the network, the output is a 1D or 2D confidence map.
The location of maximal probability from the confidence map is used as the prediction for the location of L3, allowing the extraction of the transverse slice from the CT volume.
In this section we describe in detail each step of the pipeline.

\subsection{Pre-processing}

The input 3D CT volumes are first converted into 2D Maximal Intensity Projection (MIP) images along the frontal and sagittal views, similarly to \cite{belharbi2017spotting};
however, we compute a restricted MIP for the sagittal view in order to eliminate the outer edges of the pelvis and to have a clear view of the sacrum vertebrae, which is an essential reference point for determining the position of L3 if the annotator counts the vertabrae bottom-up.
As the spine tends to be situated in the middle of the images in the majority of cases, we only compute the restricted sagittal MIP using the range [-20, 20] from the centre of the image; however, for the rare cases where the spine is not at the centre of the image, it is potentially possible to use a more elaborate image processing technique to attempt to detect the centreline of the spine and centre it.
As CT images tend to have different slice thicknesses, we normalise the pixel size of the resulting MIPs to $1\times1mm^2$ to allow consistent input to the algorithm.

Finally, we threshold the images between 100 HU and 1500 HU in order to eliminate the majority of soft tissue at the lower end of HU and minimise the effect of metal implants and artifacts above 1500 HU.
The images are then mapped to 8bit ($[-127, 127]$).
Figure \ref{fig:mip} shows an example of MIP images obtained from a 3D CT volume.

\begin{figure}[h]
    \centering
       \begin{subfigure}[b]{0.40\textwidth}
        \includegraphics[width=1\textwidth]{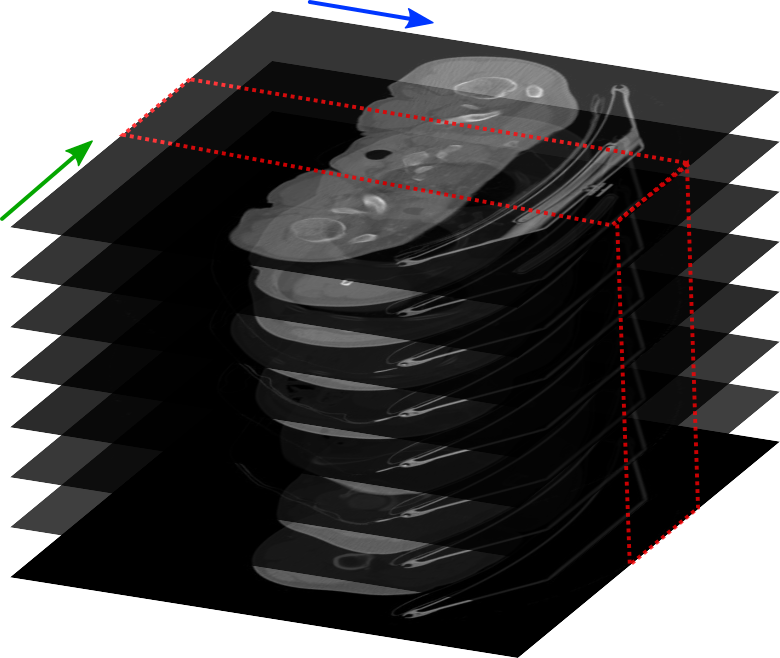}
\caption{A stack of CT slices.}
        \label{fig:mip}
    \end{subfigure}
    
        \begin{subfigure}[t]{0.31\textwidth}
        \includegraphics[width=1\textwidth]{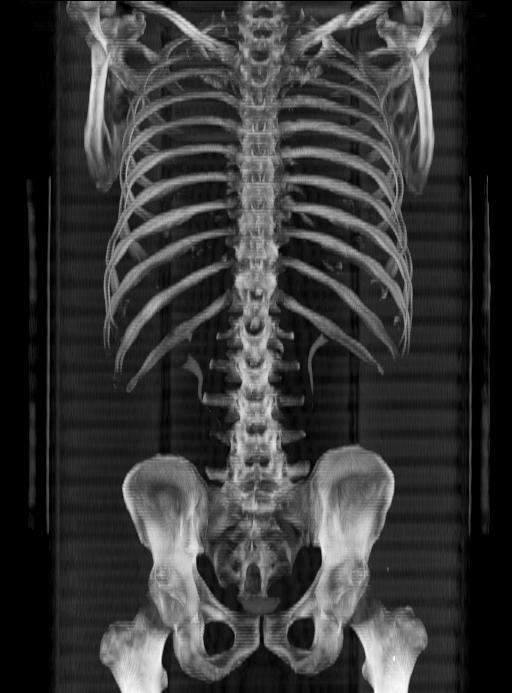}
\caption{Frontal MIP image obtained from performing a maximal intensity projection along the direction perpendicular to the frontal plane (blue in (a)).}
        \label{fig:lv}
    \end{subfigure}
    \begin{subfigure}[t]{0.31\textwidth}
        \includegraphics[width=1\textwidth]{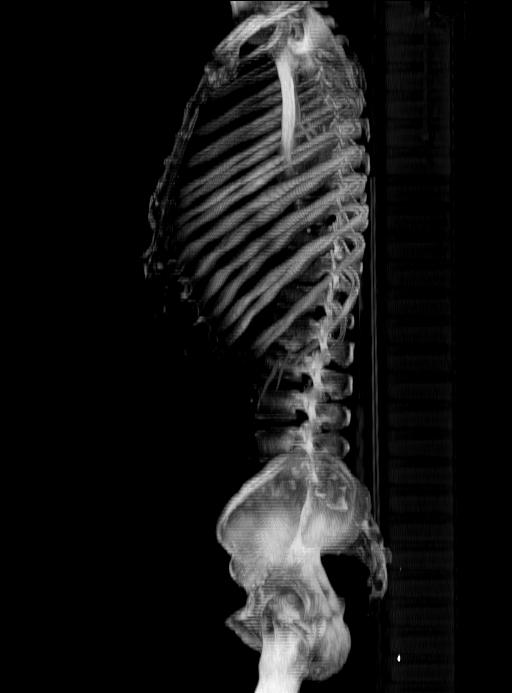}
\caption{Sagittal MIP image obtained from performing a maximal intensity projection along the direction perpendicular to the sagittal plane (green in (a)).}
        \label{fig:lv}
    \end{subfigure}
    ~ %
    \begin{subfigure}[t]{0.31\textwidth}
       \includegraphics[width=1\textwidth]{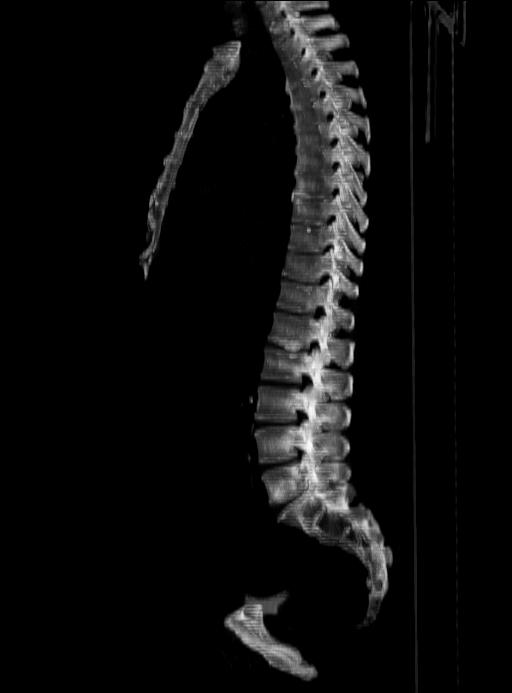}
\caption{Restricted sagittal MIP image computed only on the restricted range (in red) of the CT in (a)}
        \label{fig:rlv}
    \end{subfigure}

    \caption{Maximal intensity projection (MIP) images\label{fig:mip}}
\end{figure}

\subsection{Model Architecture}

\begin{figure}[!h]
\centering

  \begin{subfigure}[t]{1\textwidth}
     \includegraphics[width=1\textwidth]{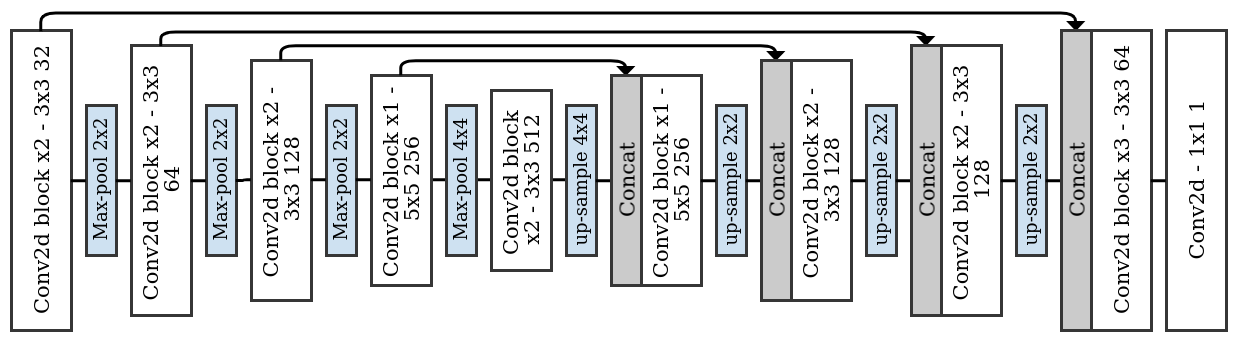}
\caption{\label{fig:FCNN2D}L3UNet-2D
}
    \end{subfigure}
    
    \vspace{0.7cm}
    
\begin{subfigure}[t]{1\textwidth}
\includegraphics[width=1\textwidth]{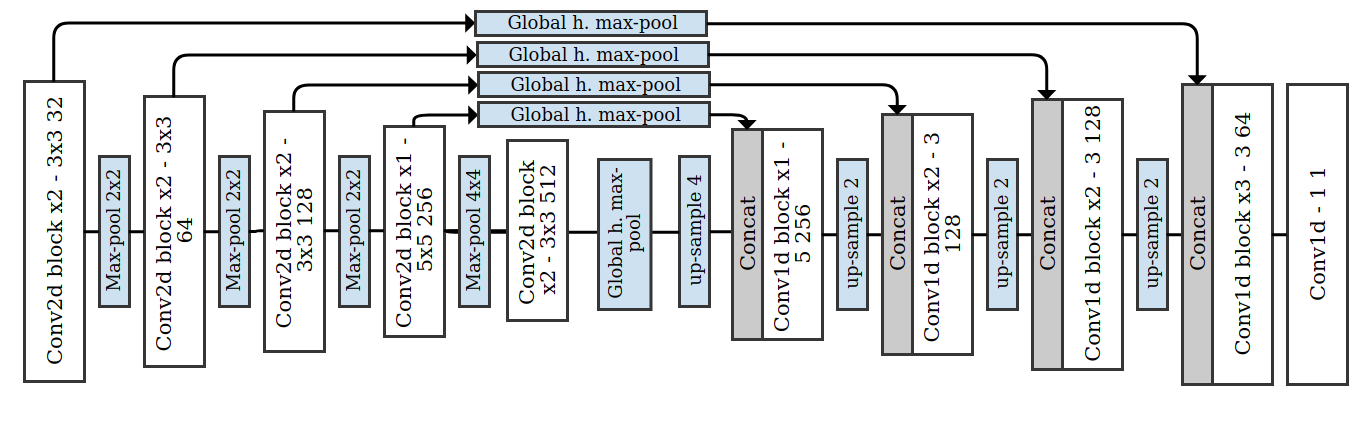}
\caption{\label{fig:FCNN1D}L3UNet-1D
}
\end{subfigure}
\caption{Network architecture. Each conv block consists of a repetition of convolutional units.
A conv unit consists of a sequence of convolution, batch normalisation, and ReLU activation.
The figure indicates each block's parameters as follows: Conv<dim> block x<number of units> - <kernel size> <number of layers>. Arrows indicate skip connections. Global h. max-pool refers to global horizontal max-pooling.}
\end{figure}

We investigate the use of a FCNN based on the UNet architecture \cite{ronneberger2015u} with a 2D confidence map output, and we propose a variant architecture that allows for a 1D confidence map output.

\paragraph{2D output} The FCNN is based on a UNet-like  architecture, and it consists of multiple down-sampling and up-sampling blocks, with the latter mirroring the former.
Each block consists of 1 or 2 convolutional units, where each unit is a sequence of 3x3 convolution, batch normalisation, and Leaky ReLU ($\alpha = 0.05$) activation.
Each block in the up-sampling path ends with an additional 1x1 convolution with the same number of channels as the corresponding block.
2x2 max-pooling is performed at the end of each block in the down-sampling path, with a 4x4 max-pooling performed at the last layer of the down-sampling path.
Skip connections, which consist of the concatenation of the output of the up-sampling blocks with the outputs from the down-sampling blocks at the same level, are used to link the down-sampling path with the up-sampling path.
In the up-sampling path we use spatial dropout at each block with $p=0.25$.
At the output of the network we attach a 1x1 convolutional layer with a sigmoid activation function and 1 channel output for the confidence map prediction.
The resulting 2D confidence map prediction output has the same dimensions as the input image.
We refer to this network as L3UNet-2D. 
Figure \ref{fig:FCNN2D} illustrates the architecture. 
The network has 8,493,537 parameters.

\paragraph{1D output} The FCNN is similarly based on the UNet architecture; the down-sampling path is the same as that of L3UNet-2D; however, the main difference is that we apply global horizontal max-pooling along the up-sampling path.
The resulting up-sampling path consists of 1D convolutions, and we employ dropout with $p=0.25$.
The output is 1D and has the same size as the height of the image.
We refer to this network as L3UNet-1D. 
Figure \ref{fig:FCNN1D} illustrates the architecture of the network. 
The network has 6,189,025 parameters.

\subsection{Augmentation}

Augmentation is a straightforward approach that typically helps in improving generalisation performance.
Image transformations are applied on the input images to create artificial variants.
We use a set of image transformations such as: horizontal flipping, scaling [0.8, 1.2], intensity offsets [-70, 70], piece-wise affine deformation, region drop-outs and over-exposures (to simulate occlusions), and vertical image sub-sampling (to simulate different slice thicknesses).
In MIP images, occlusions can show up due to the presence of metal implants, bowel content, or contrast agents;
region drop-outs and over-exposures can help make the algorithm less susceptible to such occlusions.
To simulate images with a variety of slice thicknesses (up to 7mm), an image is down-sampled along the vertical axis and then up-sampled back to its original size using linear interpolation.

\subsection{Localisation}

 We convert point-wise annotations into confidence maps $H_i$ for each MIP image $i$.
The only available annotation is the position of the slice along the y-axis.
In the frontal MIP images, the spine is expected to be located in most of the cases within a fixed range along the x-axis.
And we make this assumption when we generate the ground truth 2D confidence maps for the frontal MIP images.
Given $y_i$ as the ground truth coordinates of the L3 slice for image $i$ along the y-axis, the value of the 2D confidence map at any coordinates $(x, y)$ is defined as 
\begin{equation}
\mathbf{H}_i(x, y) = A \times (f_i*g_\sigma)(x,y),
\end{equation}

where $f_i$ is the step function

\begin{equation}
f_i(x,y)=\begin{cases} 
      1 & x_0-v\leq x\leq x_0+v, y = y_i \\
      0 & \text{otherwise},
   \end{cases}
\end{equation}

\begin{figure}
    \centering
    
    \begin{subfigure}[b]{0.23\textwidth}
        \includegraphics[width=1\textwidth]{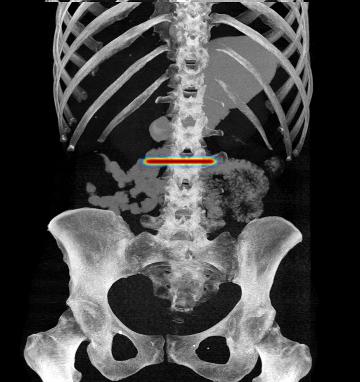}
        \caption{}
         \end{subfigure}
            \begin{subfigure}[b]{0.23\textwidth}
        \includegraphics[width=1\textwidth]{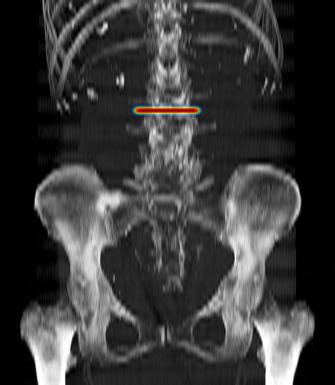}
        \caption{}
    \end{subfigure}
            \begin{subfigure}[b]{0.23\textwidth}
        \includegraphics[width=1\textwidth]{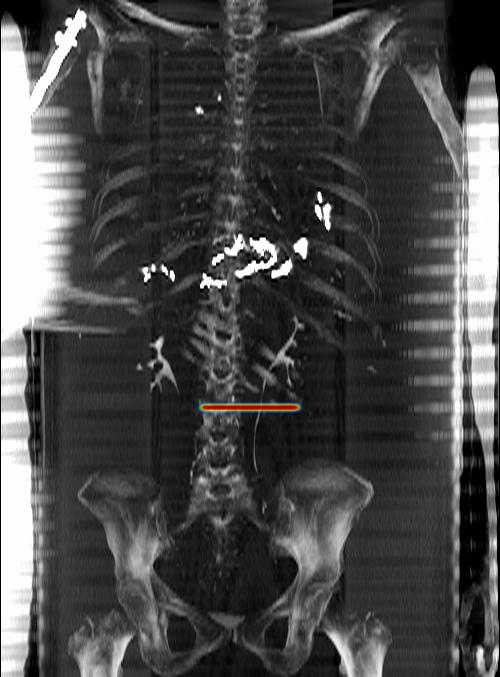}
        \caption{}
    \end{subfigure}
      \begin{subfigure}[b]{0.23\textwidth}
        \includegraphics[width=1\textwidth]{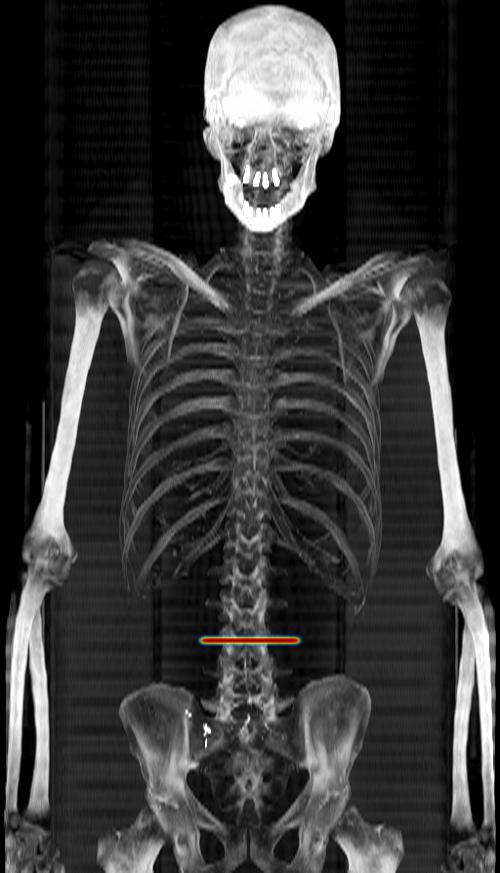}
        \caption{}
         \end{subfigure}
\caption{\label{fig:image_overlays}Examples of different images with the ground-truth confidence maps ($\sigma = 3mm$, $v=50$) overlaid on top of the images. As the only available annotation is the y coordinate, we generate an asymmetric Gaussian map centred in the middle of the image to account for offsets of the spine in the x coordinate. The dataset comprises of different images with a variety of fields of view, slice thicknesses, and artefacts.}
\end{figure}

$g_\sigma$ is a Gaussian filter function, v is an offset, and A is the max norm $||H_i(x,y)||_\infty$.
Figure \ref{fig:image_overlays} shows examples of the generated 2D confidence maps overlaid on top of the images.
In the 1D case, $f_i$ is reduced to an indicator function
\begin{equation}
f_i(y)=\begin{cases} 
      1 &  y = y_i \\
      0 & \text{otherwise}.
   \end{cases}
\end{equation}

For the objective function, we use $L_2$ loss between the predicted confidence map  $\mathbf{P}$ and the ground-truth map $\mathbf{H}$, averaged over a training batch:

\begin{align}
L = \frac{1}{N}\sum_{i=1}^N \sum_{\mathbf{x}} ||\mathbf{P}_i(\mathbf{x}) - \mathbf{H}_i(\mathbf{x})||^2,
\end{align}
where $N$ is the batch size.

\section{Experiments and Results}

We implement our architecture using keras \cite{chollet2015keras} with the tensorflow backend.
Data augmentation is performed on the fly via the data input generator using imgaug library\footnote{https://github.com/aleju/imgaug}.
We compare our proposed method with the sliding window regression approach as described in \cite{belharbi2017spotting}, as well as a modified version where we produce two outputs: the y coordinate of the slice, and a Boolean that indicates the presence or absence of the slice  (to allow training on images that do not contain a view of the L3 vertebra).

\subsection{Dataset}
\begin{figure}[!h]
    \centering
    
        \begin{subfigure}[b]{0.49\textwidth}
        \includegraphics[width=1\textwidth]{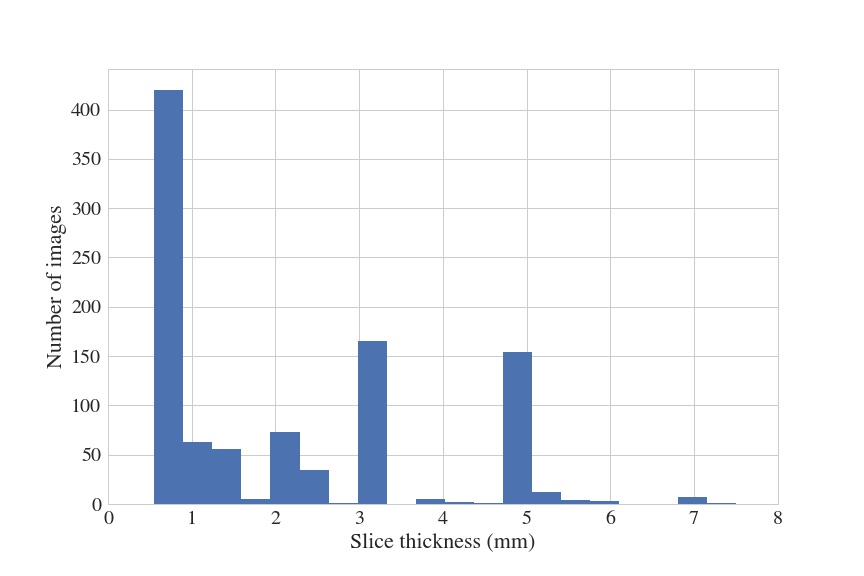}
        \caption{}
    \end{subfigure}
    \begin{subfigure}[b]{0.49\textwidth}
        \includegraphics[width=1\textwidth]{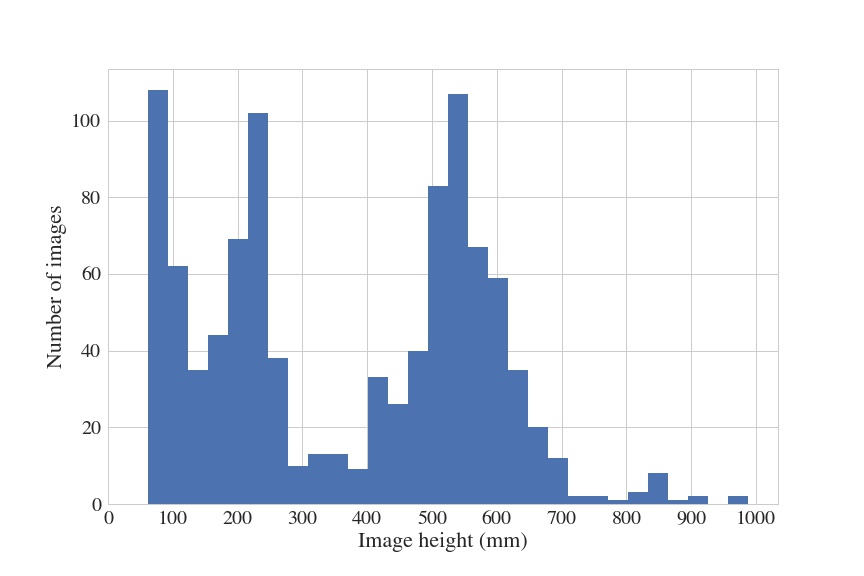}
        \caption{}

    \end{subfigure}

    \caption{Distribution of slices thicknesses (a) and image heights (b) in the dataset.}\label{fig:dataset_slice_thicknesses}
\end{figure}

We collected a diverse dataset consisting of \numImages CT images from multiple publicly available datasets.
 3 sets were obtained from the Cancer Imaging Archive (TCIA) \footnote{http://www.cancerimagingarchive.net/}: head and neck\footnote{http://doi.org/10.7937/K9/TCIA.2017.umz8dv6s}, ovarian \footnote{http://dx.doi.org/10.7937/K9/TCIA.2016.NDO1MDFQ}, colon; a liver tumour dataset is obtained from the LiTS segmentation challenge\footnote{https://competitions.codalab.org/competitions/17094}; and an ovarian cancer dataset is obtained from Hammersmith Hospital (HH), London.
Figure \ref{fig:dataset_slice_thicknesses} provides the distribution of slice thicknesses and image heights the combined dataset.

All the \numImages 3D CT images were pre-processed, where each 3D image results in a set consisting of a frontal image and a restricted sagittal image.
The annotation were carried out on images normalised to 1x1mm.
The MIP images were annotated by 2 annotators: a radiologist with 7 years of experience and an annotator with 5 years of experience working with CT images.
For each image set, the annotator was presented with the frontal and restricted sagittal MIPs side by side, and the annotator clicked on the location of the L3 slice.
The main landmark was chosen as the middle of the pedicle, lining up with the top edge of the transverse process.
Only the position along the y-axis was recorded.
It took about 2-4 seconds to manually annotate a single image.
9 images had disagreements between the annotators and \numTransImages out of \numImages were ambiguous as there was uncertainty in assigning the location of the L3 vertebra.
Further inspection by a senior radiologist with 18 years of experience as a consultant revealed that the majority of ambiguous cases consisted of patients with congenital vertebral anomalies, with the principal anomaly being transitional vertebra.
Transitional vertebrae are ones that exhibit ambiguous characteristics, they are  relatively common in the population (15-35\%) \cite{carrino2011effect, uccar2013retrospective},  and they occur at the junction between spinal segments with various degrees of apparent transition: atlanto-occipital junction, cervicothoracic junction (with a cervical rib from C7), thoracolumbar junction (with lumbar rib at L1 or a 13th rib from T13), and lumbosacral junction (commonly refered to as Lumbosacral transitional vertebrae (LSTV)).
Inaccurate identification of the correct level due to LSTV has led to procedures being carried out at the wrong vertebra level \cite{konin2010lumbosacral}.
Correct identification of L3 in ambiguous cases can only be resolved if the image contains a view of the whole spine \cite{bron2007clinical, carrino2011effect}.
Results reported in Table \ref{tab:interrater_error} correspond to the errors where annotations from both annotators fell within the same vertebra for each image.
The error in slices was computed by dividing the error in mm of a given image by the slice thickness, without rounding.
Transitional vertebra cases (\numTransImages) were excluded, leaving \numTrainImages images for the training process; however, we still evaluated the detection algorithm on the transitional cases to verify the output, as in real-world scenarios, such cases are expected to be encountered.
The average (rounded down) L3 slice location from the 2 annotators is used as ground truth for training.

\begin{table}
\centering
\begin{tabular}{|c|c|c|c|c|}
\hline
   & mean & std & median & max  \\ \hline
 Error between A and B (mm)  & 1.90  & 1.76  & 1.00 & 9.00 \\

 Error between A and B (slice)  & 1.94 & 2.36 & 0.80 & 11.43\\
 \hline
 Error between A/B and mean of A and B (mm)  & 1.14  & 0.97  & 1.00 & 5.00 \\

  Error between A/B and mean of mean A and B (slice)  & 0.97 & 1.18 & 0.40 & 5.71 \\
 \hline
\end{tabular}
\caption{\label{tab:interrater_error}Error, computed as absolute difference, between annotators A and B. Errors are reported in mm and in number of slices. As the mean slice location from both annotators is later taken as ground truth for training, we report the error between any given annotator and the ground truth.}
\end{table}

\subsection{Results}

\begin{table}[h]
\centering
\begin{tabular}{|c|c|c|c|c|c|c |}
 \hline
  & & mean & std & median & max & > 10  \\ \hline

  \multirow{2}{*}{\textbf{L3UNet-2D} - frontal}  &error (mm)  & 2.09 & 4.56 & 1.00 & 51.00 & 19 \\
   
&error (slice)   & 1.43 & 3.52 & 0.80 & 51.00 & 12 \\
\hline
  \multirow{2}{*}{\textbf{L3UNet-1D} - frontal}  &error (mm)  & 2.12 & 4.56 & 1.00 & 38.00 & 22 \\
   
&error (slice)   & 1.53 & 4.22 & 0.67 & 45.71 & 15 \\
\hline
  \multirow{2}{*}{\textbf{L3UNet-1D} - sagittal}  &error (mm) & 1.99 & 5.41 & 1.00 & 52.00 & 28
 \\
   
&error (slice)   & 1.41 & 5.02 & \textbf{0.50} & 65.00 & 23 \\
\hline

 \multirow{2}{*}{VGG16 Regression \cite{belharbi2017spotting} - frontal}  &error (mm)  & 13.78 & 8.57 & 12.00 & 48.00 & 591 \\
   
&error (slice)  & 10.26 & 9.92 & 6.12 & 60.00 & 360 \\
 \hline
 \multirow{2}{*}{VGG16 Regression with dual output - frontal}  &error (mm)  & 6.94 & 5.90 & 6.00 & 62.00 & 191 \\
   
&error (slice)  & 5.54 & 6.29 & 3.20 & 40.00 & 180 \\
 \hline
\end{tabular}
\caption{\label{tab:crossval_error}3-fold cross validation results using our method compared to the method from \cite{belharbi2017spotting} on our dataset.
The error in slices is computed as the error in mm divided by the slice thickness for a given image, without rounding.
We also report the number of outlier images that have an error greater than 10.
}
\end{table}

For L3UNet-1D and 2D, we trained using image crops of size [256, 384], with crops randomly sampled along the y-axis and centred along the x-axis.
We used a batch size of $5$ for L3UNet-2D and $8$ for L3UNet-1D. 
We set $\sigma=1.5$ for generating the confidence maps.
During training we found it helpful to start with a larger $\sigma=10$ and linearly reduce it to 1.5 as the training progresses.
The networks were trained for 50 epochs (sufficient enough to allow no further improvements on a small validation subset of the training set) using the Adam optimiser with a learning rate of 1e-3.
For L3UNet-2D we only used the frontal MIP image as input
During testing, the whole MIP image was provided as input to the network. 
The image were padded as necessary to ensure that the height and width were divisible by the amount of max-pooling in the network.

For the sliding window VGG16 regression, we attempted to reproduce the methodology as described in \cite{belharbi2017spotting} using their suggested parameters; however, not all parameters needed for training were reported; to fine-tune the network, we use a smaller learning rate $1e-5$ to avoid quickly destroying the learned weights of VGG16, and we used a batch size of 12.
We used crop windows of size [100, 512], as recommended, sampled from frontal MIP image regions that contain the L3 vertebrae.
In addition, we trained a modified version of VGG16 regression with dual output: a y coordinate and a Boolean that indicates the presence or absence of the L3 vertebra. 
This allowed training on positive and negative image crops.
We report the results of 3-fold cross validation in Table  \ref{tab:crossval_error}.

Training and testing were carried out on a workstation with 2 Nvidia TitanX GPUs.
Once the MIP images were obtained, the average testing times for an image with average 440mm height were: L3UNet-1D 0.06s, L3UNet-2D 0.17s, sliding window VGG16 2.98s.

\begin{figure}
    \centering
     \begin{subfigure}[b]{0.23\textwidth}
        \includegraphics[width=1\textwidth]{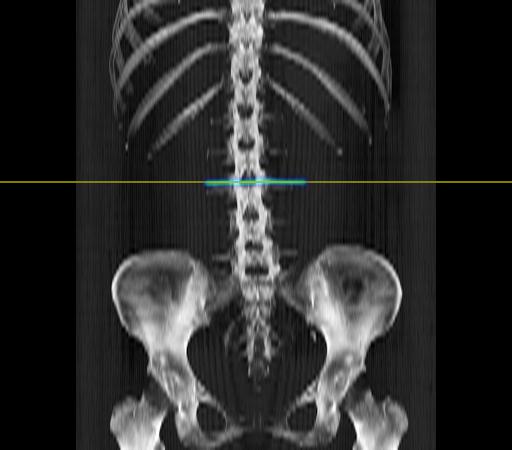}
        \caption{L3UNet-2D\\ frontal}
         \end{subfigure}
        \begin{subfigure}[b]{0.23\textwidth}
        \includegraphics[width=1\textwidth]{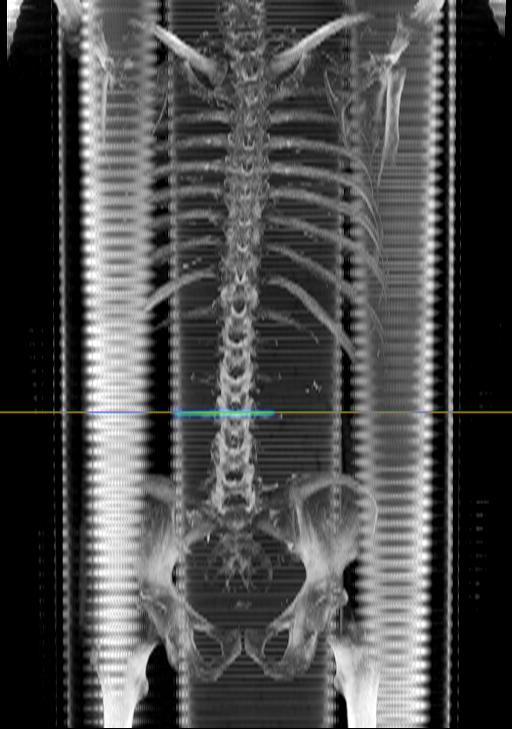}
        \caption{L3UNet-2D\\ frontal}
    \end{subfigure}
    \begin{subfigure}[b]{0.23\textwidth}
        \includegraphics[width=1\textwidth]{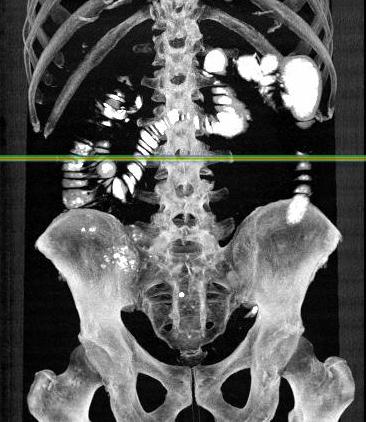}
        \caption{L3UNet-1D \\frontal}
         \end{subfigure}
            \begin{subfigure}[b]{0.23\textwidth}
        \includegraphics[width=1\textwidth]{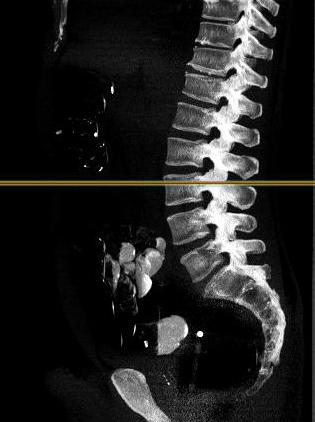}
        \caption{L3UNet-1D \\sagittal}
     
    \end{subfigure}
\caption{\label{fig:example_output}Examples of prediction output. The predicted confidence map is overlaid on the images. For the 1D case, the confidence map is stretched out along the x-axis for visualisation purposes. The red line corresponds to prediction, while green line is the ground truth.}
\end{figure}

\begin{figure}[h]
\centering
\includegraphics[width=0.5\textwidth]{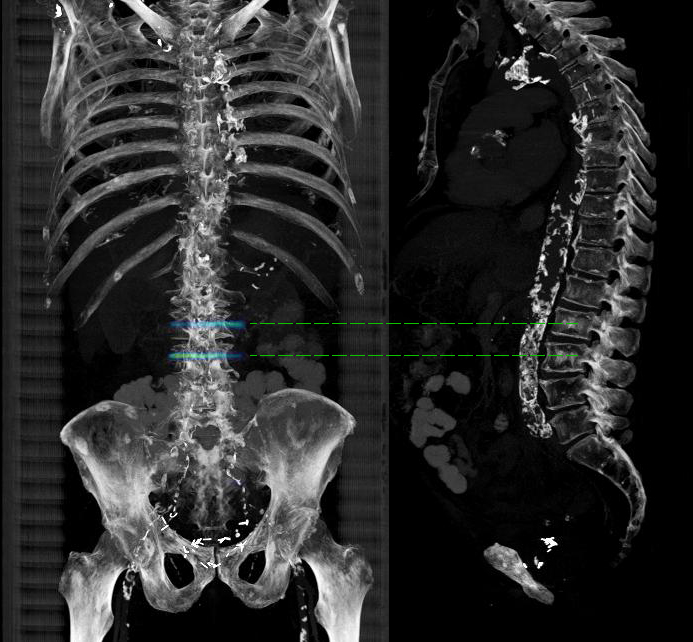}
 
\caption{\label{fig:transitional} Output on a transitional vertebra case by L3UNet-2D. The confidence map indicates two potential locations for L3.   }
\end{figure}

Finally, we applied our trained L3UNets on the excluded set of transitional vertebra cases.
Outputs in all cases consisted in one of the two adjacent true candidate vertebrae or, in some cases, both. 
Figure \ref{fig:transitional} show a sample output on a transitional vertebra case.

\subsection{Discussion}

Results show that FCNNs with confidence map outputs achieve state-of-the-art L3 localisation performance, with a median error using our proposed L3UNet models equal to 1mm, similar to the human annotator error.
We note that using our proposed approach of 1D confidence map outputs achieves slightly better results than the 2D confidence map output, despite the global max-pooling applied along the up-sampling path.
As the only available annotation is the slice location along the y-axis, it is straight-forward to apply L3UNet-1D on either the frontal or sagittal MIP images.
However, with the 2D output, we only make use of the frontal MIP image as we make the assumption that the spine is located in the centre of the image.
In the sagittal MIP images, the spine is not always located at the same position.
Although, it would have been possible to potentially detect the spine with some image processing, we have not done so.
Therefore, we only use the frontal image as input for L3UNet-2D.

\begin{figure}[h]

    \centering
    
        \begin{subfigure}[t]{0.48\textwidth}
        \includegraphics[width=1\textwidth]{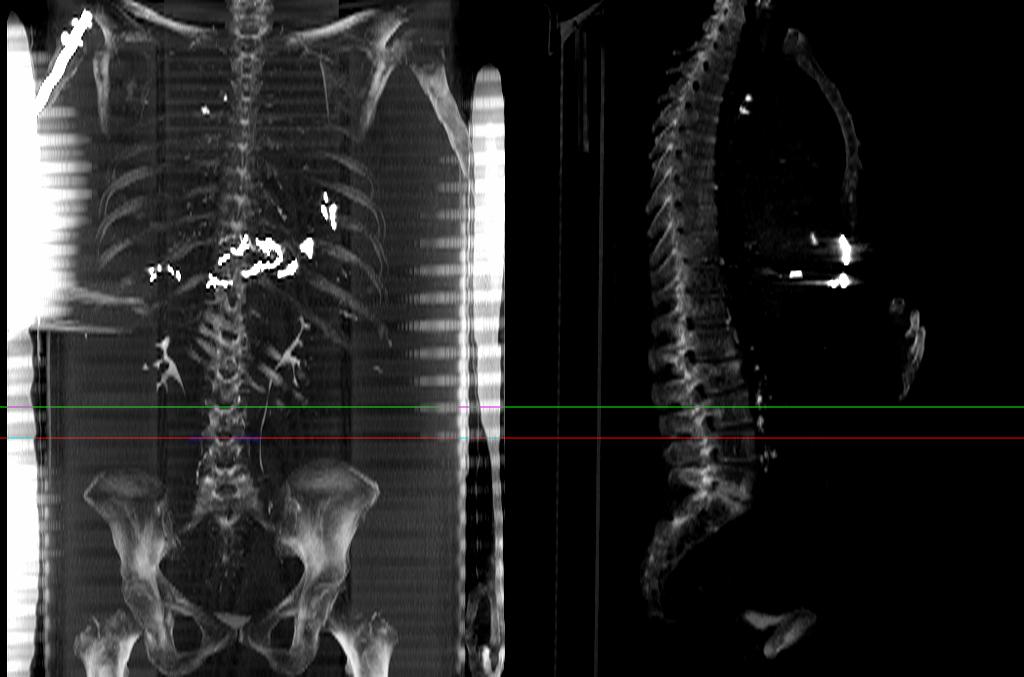}
        \caption{Incorrect identification of the L3 level by L3UNet-2D potentially due to the low contrast of the spine, rendering the transverse process almost invisible. The outputted confidence map is low.}
    \end{subfigure}
   \quad
            \begin{subfigure}[t]{0.48\textwidth}
        \includegraphics[width=1\textwidth]{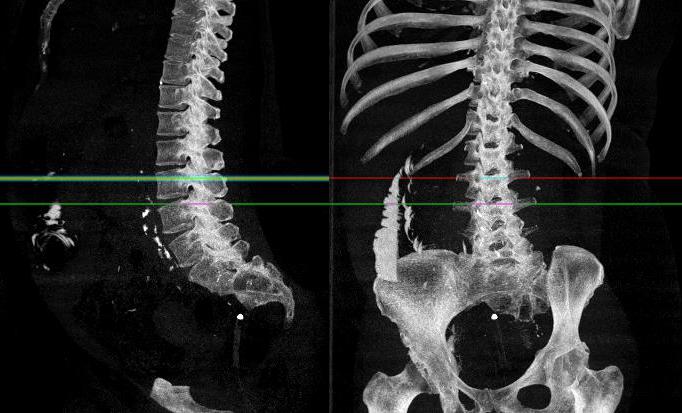}
        \caption{Incorrect identification of the L3 level by L3UNet-1D with sagittal image input. The spine is curved, resulting in an apparent merging of L5 with the sacrum.}

    \end{subfigure}
\caption{\label{fig:outliers} Examples of incorrect identification of outlier cases.}
\end{figure}

There still remains a small subset of images where the prediction is maximally off by one vertebra (outliers are images with an error more than 10mm).
An inspection of the outlier images revealed that roughly half of them have an apparent reason that might explain why the network made an incorrect prediction. Figure \ref{fig:outliers} shows such an example of outliers.
In addition, a few outliers had probabilities less than 50\%, so it would potentially be possible to detect them.
Nonetheless, the other half of outliers had confident, incorrect prediction outputs with no apparent reason.
It is potentially likely, though, despite the best efforts of the annotators that some transitional vertebrae cases have slipped past them into the training set, given that the cases are at most prevalent in the general population 15-35\%, while only \numTransImages out of \numImages, which amounts to about 5\% were found in our dataset.
With transitional vertebrae cases, it is not possible to determine the correct L3 level without a full view of the spine, which would allow counting from the cervical segment.
We tested our network on the excluded set of transitional vertebrae images, the outputted confidence maps always resulted in predictions for one or the other potential L3 candidate, and occasionally both.
In a practical application where the goal is performing sarcopenia measurement, 
it would be worth investigating the amount of tolerable offset from L3 that would result in a different prognostic output; this is of course also subject to accurate segmentation of the muscle area.

The results that we obtained using the sliding window approach of VGG16 regression are  not within the same range as those reported in \cite{belharbi2017spotting} despite attempts to improve the output.
This could be due to three things: (1) our dataset is more complex than the one used in \cite{belharbi2017spotting} and contains more finer slice thicknesses (minimum of 0.8mm vs 2mm); however, their dataset is not publicly available, so there is no means to verify this. (2) Based on previous reported results in the literature  (\cite{payer2016regressing, tompson2015efficient}) on the use of networks with one down-sampling path, there is always a reduction in localisation accuracy due to max-pooling: this is the case with VGG16, so the obtained results are somewhat to be expected. 
In addition \cite{belharbi2017spotting} report that vertical max-pooling
distorts the target position when using a custom network trained from scratch, but this does not seem to be reflected in the results reported for VGG16. (3) An implementation error of their slice detection method on our part, despite our best efforts of reproduction.
Nonetheless, disregarding the accuracy results, one clear advantage of our proposed method is the the inference speed, where 50x speed-up can be obtained using L3UNet-1D.
Using either L3UNet-1D or 2D, a prediction can be made in 0.06s and 0.17s, respectively, for an image of average height 440mm.

\section{Conclusion}

We have proposed an efficient method for the automatic detection of the L3 slice within a 3D CT volume, with the principal goal of using our method as part of an automatic sarcopenia measurement tool.
The method is based on a fully-convolutional UNet-like architecture with no restriction on the image input size.
A 3D CT image is first converted into a 2D MIP image, reducing the dimensionality of the problem.
MIP images and associated confidences maps are then used to train a FCNN.
We have in addition proposed a novel architecture variant that consists in a simple modification of the UNet architecture that is suited for our particular use case of predicting a confidence map along a single axis.
Our method achieves state-of-the-art results with a median error rate (in mm) that is comparable to the inter-annotator median error.
Future work could look at the potential of using our method for multiple vertebrae detection, instead of a single vertebra, and comparing that with 3D methods.
3D coordinates of a vertebra could be obtained using the frontal and sagittal MIP images, and the traverse slices. 
The algorithm could first detect the coordinates of the vertabrae along the sagittal and frontal planes, and then along the transverse planes.

\section*{Acknowledgements}
This work was supported by United Kingdom NIHR Biomedical Research Centre award to Imperial College London. 
We acknowledge programmatic support from Imperial College Experimental Cancer Medicines Centres and United Kingdom Medical Research Council (MR/N020782/1)

\bibliographystyle{alpha}

\begin{thebibliography}{TVVR{\etalchar{+}}15}

\bibitem[BCH{\etalchar{+}}17]{belharbi2017spotting}
Soufiane Belharbi, Cl{\'e}ment Chatelain, Romain H{\'e}rault, S{\'e}bastien
  Adam, S{\'e}bastien Thureau, Mathieu Chastan, and Romain Modzelewski.
\newblock Spotting l3 slice in ct scans using deep convolutional network and
  transfer learning.
\newblock {\em Computers in biology and medicine}, 87:95--103, 2017.

\bibitem[Bro07]{bron2007clinical}
Johannes~L Bron.
\newblock The clinical significance of lumbosacral transitional anomalies.
\newblock {\em Acta Orthopaedica Belgica}, 73(6):687, 2007.

\bibitem[C{\etalchar{+}}15]{chollet2015keras}
Fran\c{c}ois Chollet et~al.
\newblock Keras.
\newblock \url{https://keras.io}, 2015.

\bibitem[CCJL{\etalchar{+}}11]{carrino2011effect}
John~A Carrino, Paul~D Campbell~Jr, Dennis~C Lin, William~B Morrison, Mark~E
  Schweitzer, Adam~E Flanders, John Eng, and Alexander~R Vaccaro.
\newblock Effect of spinal segment variants on numbering vertebral levels at
  lumbar mr imaging.
\newblock {\em Radiology}, 259(1):196--202, 2011.

\bibitem[CSQ{\etalchar{+}}15]{chen2015automatic}
Hao Chen, Chiyao Shen, Jing Qin, Dong Ni, Lin Shi, Jack~CY Cheng, and Pheng-Ann
  Heng.
\newblock Automatic localization and identification of vertebrae in spine ct
  via a joint learning model with deep neural networks.
\newblock In {\em International Conference on Medical Image Computing and
  Computer-Assisted Intervention}, pages 515--522. Springer, 2015.

\bibitem[CSWS17]{cao2017realtime}
Zhe Cao, Tomas Simon, Shih-En Wei, and Yaser Sheikh.
\newblock Realtime multi-person 2d pose estimation using part affinity fields.
\newblock In {\em CVPR}, 2017.

\bibitem[GRCD11]{ghosh2011automatic}
Subarna Ghosh, Alomari Raja'S, Vipin Chaudhary, and Gurmeet Dhillon.
\newblock Automatic lumbar vertebra segmentation from clinical ct for wedge
  compression fracture diagnosis.
\newblock In {\em Medical Imaging 2011: Computer-Aided Diagnosis}, volume 7963,
  page 796303. International Society for Optics and Photonics, 2011.

\bibitem[GZK{\etalchar{+}}13]{glocker2013vertebrae}
Ben Glocker, Darko Zikic, Ender Konukoglu, David~R Haynor, and Antonio
  Criminisi.
\newblock Vertebrae localization in pathological spine ct via dense
  classification from sparse annotations.
\newblock In {\em International Conference on Medical Image Computing and
  Computer-Assisted Intervention}, pages 262--270. Springer, 2013.

\bibitem[KBMB16]{kazemi2016computed}
Seyyed Mohammad~Reza Kazemi-Bajestani, Vera~C Mazurak, and Vickie Baracos.
\newblock Computed tomography-defined muscle and fat wasting are associated
  with cancer clinical outcomes.
\newblock In {\em Seminars in cell \& developmental biology}, volume~54, pages
  2--10. Elsevier, 2016.

\bibitem[KIL{\etalchar{+}}15]{korez2015framework}
Robert Korez, Bulat Ibragimov, Bo{\v{s}}tjan Likar, Franjo Pernu{\v{s}}, and
  Toma{\v{z}} Vrtovec.
\newblock A framework for automated spine and vertebrae interpolation-based
  detection and model-based segmentation.
\newblock {\em IEEE transactions on medical imaging}, 34(8):1649--1662, 2015.

\bibitem[KW10]{konin2010lumbosacral}
GP~Konin and DM~Walz.
\newblock Lumbosacral transitional vertebrae: classification, imaging findings,
  and clinical relevance.
\newblock {\em American Journal of Neuroradiology}, 31(10):1778--1786, 2010.

\bibitem[LHC{\etalchar{+}}15]{lerouge2015ioda}
Julien Lerouge, Romain Herault, Cl{\'e}ment Chatelain, Fabrice Jardin, and
  Romain Modzelewski.
\newblock Ioda: an input/output deep architecture for image labeling.
\newblock {\em Pattern recognition}, 48(9):2847--2858, 2015.

\bibitem[LTT{\etalchar{+}}17]{lee2017pixel}
Hyunkwang Lee, Fabian~M Troschel, Shahein Tajmir, Georg Fuchs, Julia Mario,
  Florian~J Fintelmann, and Synho Do.
\newblock Pixel-level deep segmentation: artificial intelligence quantifies
  muscle on computed tomography for body morphometric analysis.
\newblock {\em Journal of digital imaging}, 30(4):487--498, 2017.

\bibitem[MCB{\etalchar{+}}12]{mir2012sarcopenia}
Olivier Mir, Romain Coriat, Benoit Blanchet, Jean-Philippe Durand, Pascaline
  Boudou-Rouquette, Judith Michels, Stanislas Ropert, Michel Vidal, Stanislas
  Pol, Stanislas Chaussade, et~al.
\newblock Sarcopenia predicts early dose-limiting toxicities and
  pharmacokinetics of sorafenib in patients with hepatocellular carcinoma.
\newblock {\em PloS one}, 7(5):e37563, 2012.

\bibitem[PCZ15]{pfister2015flowing}
Tomas Pfister, James Charles, and Andrew Zisserman.
\newblock Flowing convnets for human pose estimation in videos.
\newblock In {\em Proceedings of the IEEE International Conference on Computer
  Vision}, pages 1913--1921, 2015.

\bibitem[P{\v{S}}BU16]{payer2016regressing}
Christian Payer, Darko {\v{S}}tern, Horst Bischof, and Martin Urschler.
\newblock Regressing heatmaps for multiple landmark localization using cnns.
\newblock In {\em International Conference on Medical Image Computing and
  Computer-Assisted Intervention}, pages 230--238. Springer, 2016.

\bibitem[RFB15]{ronneberger2015u}
Olaf Ronneberger, Philipp Fischer, and Thomas Brox.
\newblock U-net: Convolutional networks for biomedical image segmentation.
\newblock In {\em International Conference on Medical image computing and
  computer-assisted intervention}, pages 234--241. Springer, 2015.

\bibitem[RRA13]{rasoulian2013lumbar}
Abtin Rasoulian, Robert Rohling, and Purang Abolmaesumi.
\newblock Lumbar spine segmentation using a statistical multi-vertebrae
  anatomical shape+ pose model.
\newblock {\em IEEE transactions on medical imaging}, 32(10):1890--1900, 2013.

\bibitem[SSL{\etalchar{+}}15]{suzani2015fast}
Amin Suzani, Alexander Seitel, Yuan Liu, Sidney Fels, Robert~N Rohling, and
  Purang Abolmaesumi.
\newblock Fast automatic vertebrae detection and localization in pathological
  ct scans-a deep learning approach.
\newblock In {\em International Conference on Medical Image Computing and
  Computer-Assisted Intervention}, pages 678--686. Springer, 2015.

\bibitem[SWMN16]{shachar2016prognostic}
Shlomit~Strulov Shachar, Grant~R Williams, Hyman~B Muss, and Tomohiro~F
  Nishijima.
\newblock Prognostic value of sarcopenia in adults with solid tumours: a
  meta-analysis and systematic review.
\newblock {\em European journal of cancer}, 57:58--67, 2016.

\bibitem[TGJ{\etalchar{+}}15]{tompson2015efficient}
Jonathan Tompson, Ross Goroshin, Arjun Jain, Yann LeCun, and Christoph Bregler.
\newblock Efficient object localization using convolutional networks.
\newblock In {\em Proceedings of the IEEE Conference on Computer Vision and
  Pattern Recognition}, pages 648--656, 2015.

\bibitem[TVVR{\etalchar{+}}15]{tegels2015sarcopenia}
Juul~JW Tegels, Jeroen~La Van~Vugt, Kostan~W Reisinger, Karel~WE Hulsew{\'e},
  Anton~GM Hoofwijk, Joep~PM Derikx, and Jan~HMB Stoot.
\newblock Sarcopenia is highly prevalent in patients undergoing surgery for
  gastric cancer but not associated with worse outcomes.
\newblock {\em Journal of surgical oncology}, 112(4):403--407, 2015.

\bibitem[UUC{\etalchar{+}}13]{uccar2013retrospective}
Demet U{\c{c}}ar, Bekir~Yavuz U{\c{c}}ar, Yahya Co{\c{s}}ar, Kurtulu{\c{s}}
  Emrem, G{\"u}rkan G{\"u}m{\"u}{\c{s}}suyu, Serhat Mutlu, Burcu Mutlu,
  Mehmet~Akif {\c{C}}a{\c{c}}an, Y{\i}lmaz Mertsoy, and Hatice
  G{\"u}m{\"u}{\c{s}}.
\newblock Retrospective cohort study of the prevalence of lumbosacral
  transitional vertebra in a wide and well-represented population.
\newblock {\em Arthritis}, 2013, 2013.

\bibitem[YXX{\etalchar{+}}17a]{yang2017automatic}
Dong Yang, Tao Xiong, Daguang Xu, Qiangui Huang, David Liu, S~Kevin Zhou,
  Zhoubing Xu, JinHyeong Park, Mingqing Chen, Trac~D Tran, et~al.
\newblock Automatic vertebra labeling in large-scale 3d ct using deep
  image-to-image network with message passing and sparsity regularization.
\newblock In {\em International Conference on Information Processing in Medical
  Imaging}, pages 633--644. Springer, 2017.

\bibitem[YXX{\etalchar{+}}17b]{yang2017deep}
Dong Yang, Tao Xiong, Daguang Xu, S~Kevin Zhou, Zhoubing Xu, Mingqing Chen,
  JinHyeong Park, Sasa Grbic, Trac~D Tran, Sang~Peter Chin, et~al.
\newblock Deep image-to-image recurrent network with shape basis learning for
  automatic vertebra labeling in large-scale 3d ct volumes.
\newblock In {\em International Conference on Medical Image Computing and
  Computer-Assisted Intervention}, pages 498--506. Springer, 2017.

\end{thebibliography}
\newcommand{\etalchar}[1]{$^{#1}$}

\end{document}